\documentclass[letterpaper, 10 pt, conference]{ieeeconf}  

\IEEEoverridecommandlockouts                              

\usepackage{cite}
\usepackage{graphics} 
\usepackage{epsfig} 
\usepackage{epstopdf}
\graphicspath {{./image/}}
\usepackage{mathptmx} 
\usepackage{times} 
\usepackage{amsmath} 
\usepackage{amssymb}  
\usepackage{gensymb}
\usepackage{multirow}
\usepackage{picinpar}
\usepackage{multicol}
\usepackage{stfloats}
\usepackage{caption}
\usepackage{graphicx}
\usepackage{subfigure}
\usepackage{wasysym}
\usepackage{textcomp}
\usepackage{siunitx}
\let\labelindent\relax
\usepackage{enumitem}
\usepackage{comment}
\usepackage[colorlinks,linkcolor=red,anchorcolor=blue,citecolor=green]{hyperref}
\def\BibTeX{{\rm B\kern-.05em{\sc i\kern-.025em b}\kern-.08em
    T\kern-.1667em\lower.7ex\hbox{E}\kern-.125emX}}
    
\usepackage{xcolor}

\title{\LARGE \bf
DeepPCO: End-to-End Point Cloud Odometry through Deep Parallel Neural Network
}

\author{Wei Wang, Muhamad Risqi U. Saputra, Peijun Zhao, Pedro Gusmao, \\
Bo Yang, Changhao Chen, Andrew Markham, and Niki Trigoni
\thanks{*The authors are with the Department of Computer Science,
University of Oxford, Oxford OX1 3QD, United Kingdom.
\{firstname.lastname\}@cs.ox.ac.uk}}

\begin{document}

\maketitle
\thispagestyle{empty}
\pagestyle{empty}

\begin{abstract}
Odometry is of key importance for localization in the absence of a map. There is considerable work in the area of visual odometry (VO), and recent advances in deep learning have brought novel approaches to VO, which directly learn salient features from raw images. These learning-based approaches have led to more accurate and robust VO systems. However, they have not been well applied to point cloud data yet. In this work, we investigate how to exploit deep learning to estimate point cloud odometry (PCO), which may serve as a critical component in point cloud-based downstream tasks or learning-based systems. Specifically, we propose a novel end-to-end deep parallel neural network called DeepPCO, which can estimate the 6-DOF poses using consecutive point clouds. It consists of two parallel sub-networks to estimate 3-D translation and orientation respectively rather than a single neural network. We validate our approach on KITTI Visual Odometry/SLAM benchmark dataset with different baselines. Experiments demonstrate that the proposed approach achieves good performance in terms of pose accuracy.
\end{abstract}

\section{Introduction}

Visual odometry (VO) estimation is one of the most fundamental research tasks in the field of computer vision and robotics. It incrementally estimates an agent's relative pose by examining changes in projected geometrical features captured from a monocular camera. Conventional visual odometry relies on feature extraction and matching which has been well studied for a number of decades. However, recent advances in deep learning bring another paradigm to VO by directly inferring the 6 Degree-of-Freedom (6-DoF) camera poses through end-to-end learning \cite{Saputra2018}. This leads to several advantages in terms of not requiring hand-crafted features and being able to learn directly from large datasets. Deep learning-based VO approaches achieve very impressive results compared to conventional geometry-based approaches in some benchmark datasets \cite{Wang2017}.

Point cloud data, such as that generated by a Lidar or Depth Camera, can provide richer information about the 3-D structure of the environment. This intuitively provides a better perspective on how the sensor moves over time.  Conventional point-cloud based odometry relies on geometric approaches and has shown excellent performance~\cite{Zhang2014}. However, these techniques suffer from issues of robustness such as being sensitive to outliers and having low adaptability to different environments~\cite{Dossier2016}. Scan matching, which is a fundamental task in point-cloud based odometry, is prone to outliers introduced by hardware failure, agent vibration when collecting data, or unexpected sensor movements. Moreover, when the environment is featureless or objects within the environment are deformable, scan matching may also fail. These systems usually need other sensors such as the Inertial Measurement Unit (IMU) to compensate for these failures~\cite{Zhang2015}. Another complexity is that operators need to perform manual hand-engineering to fine-tune the large number of model parameters. 

\begin{figure}
  \centering
  \includegraphics[width=\columnwidth]{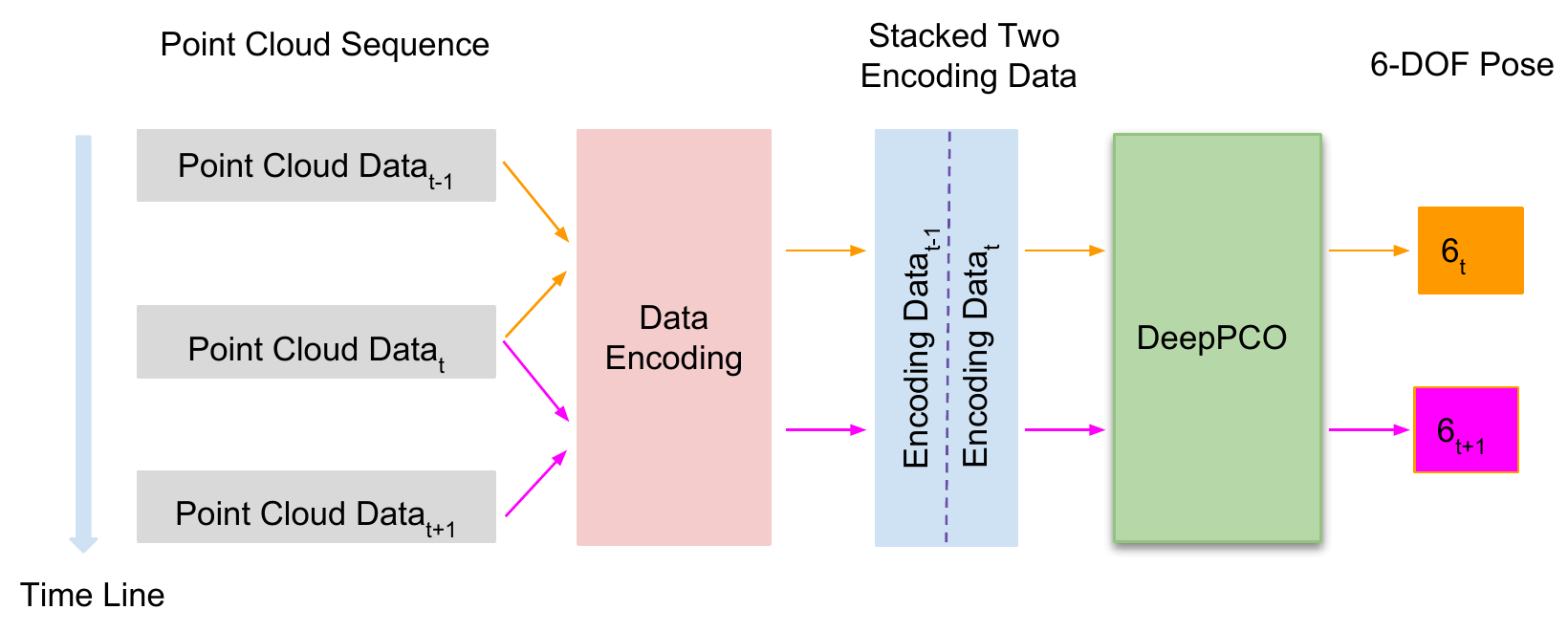}
  \caption{System overview of the proposed DeepPCO framework. A sequential point cloud data stream is sent to the system. Two consecutive point cloud data are sent in parallel to the data encoding module, which implements the 2-D panoramic projection method. Projected depth images are then stacked to feed to DeepPCO neural network for predicting 6-DOF pose. These relative transformations are further combined to form a moving trajectory.}
  \label{system_overview}
\end{figure}

Motivated by the success of deep learning as applied to VO, we consider whether similar successes can be achieved in estimating odometry from point cloud data, obviating the need for manually engineering features and model parameters, which we define as the Point Cloud Odometry (PCO) task. In particular, deep learning may eliminate the problem of noisy scan matching, potentially improving robustness. Furthermore, solving point clouds 6-DOF pose is significant for point cloud-related tasks in the realm of 3-D vision. Additionally, it is a natural fit to integrate learning-based odometry components into learning-based systems or SLAM. Nevertheless, a challenging problem for using point clouds is that it is unordered, which poses issues for convolutional kernels which inherently assume a structured input.

In this paper, we investigate two key issues which impact the performance of PCO using end-to-end deep learning. The first is the representation (encoding) of the point cloud itself. This ranges from the raw point cloud to various approaches to projecting it to a 2-D scene. The second issue comes from the choice of the model architecture itself. In particular, we consider estimating the 6-DOF pose (translation and orientation) in a single regression network, or splitting the translation and orientation estimation tasks and regressing them in two sub-networks. 

Based on these investigations, we propose a novel approach, termed DeepPCO, which combines the use of 2-D panoramic depth projections with two sub-networks to achieve accurate odometry, as shown in Fig.\ref{system_overview}. DeepPCO eliminates the intermediate modules (e.g. scan matching, geometric estimation) of a classical pipeline. We compare against various baselines using point cloud data from the KITTI Vision Benchmark Suite \cite{Geiger2012} which were collected using a \ang{360} Velodyne laser scanner.

Our main contributions are as follows:
\begin{itemize}
  \item We demonstrate that point cloud odometry problem can be effectively solved in an end-to-end fashion, and our proposed architecture outperforms existing learning-based approaches by a significant margin, and received comparable performance to conventional ones.
  \item We adopt a dual-branch architecture to infer 3-D translation and orientation separately instead of a single network.
  \item Comprehensive experiments and ablation studies have been done to evaluate our proposed method. Results show that DeepPCO achieves good performance with respect to different kinds of neural network architectures.
\end{itemize}

\section{Related Work}
In this section, we review deep learning for odometry tasks, especially Visual Odometry (VO), followed by the introduction of geometric and learning-based approaches for point cloud odometry. Previous works mainly focus on point cloud data generated by Lidar sensors, so we call this kind of work as Lidar Odometry (LO).

\subsection{Deep Learning for Visual Odometry}
Instead of manually estimating the geometry of the scenes, learning-based systems automatically learn the feature correspondences and relationships between images. These types of systems are usually trained in an end-to-end manner and save great efforts in engineering compared to the classic ones. Wang et al.~\cite{Wang2017} proposed an end-to-end framework for monocular VO consisting of a CNN branch built upon FlowNet~\cite{Fischer2015}, followed by a Recurrent Convolutional Neural Network structure (RCNN).  Features are first extracted for two consecutive images through CNN and then forwarded to LSTM. Clark et al.~\cite{Clark2017} treated visual-inertial odometry as a sequence-to-sequence problem and made use of LSTM to embed IMU information to predict pose. Chen et al.~\cite{Chen2019} proposed an selective sensor fusion approach to solve visual-inertial odometry task. Constante et al.~\cite{Costante2018} presented a novel motion estimation network called LS-VO, which is based on Auto-Encoder (AE) scheme to find a non-linear representation of the Optical Flow (OF) manifold. \cite{Muller2017} used raw optical flow images as inputs and proposed a VO system called Flowdometry, whose architecture consists of CNN. Although the translation and rotation error evaluation does not excel the state-of-the-art approaches, due to careful engineering design, it achieved 23.796x speedup over existing learning-based VO. Gomez-Ojeda et al~\cite{Gomez-Ojeda2018} developed a learning-based image enhancement approach to solve the robustness problem in VO. The network architecture consists of CNN and LSTM, and the LSTM was proved to help reduce noises by incorporating the temporal information from past sequences. Li et al.~\cite{Li2018} presented UnDeepVO to estimate the 6-DOF poses of a monocular camera and the depth of its view. Stereo image pairs were harnessed for recovering absolute scale and loss function was defined on spatial and temporal dense information. Yang et al.~\cite{Yang2018} discussed a novel approach called Deep Virtual Stereo Odometry (DVSO), which first predicts depth using monocular images, and then incorporate them into Direct Sparse Odometry (DSO) as direct virtual stereo measurements. All the mentioned works take 2-D inputs to their neural networks.

\subsection{Geometry and Deep Learning for Lidar Odometry}
Classical geometry-based LO usually employs variations of Iterative Closest Point (ICP)~\cite{Besl1992} for scan matching. \cite{Segal2009} introduced a probabilistic framework combining ICP and `point-to-plane' algorithms to model locally planar surface structure from both scans, which can be applied to the registration of frame-to-frame scans. The state-of-the-art LO system, named LOAM, was proposed by~\cite{Zhang2014}, which consists of feature point extraction, feature correspondence calculation, motion estimation, and mapping components. The key idea is to optimize a large number of variables via two algorithms. One algorithm employs low fidelity to estimate velocity and the other one performs very low frequency for fine matching and point cloud registration. These two stages can be concluded as fast scan-to-scan and precise scan-to-map. Nicolai et al.~\cite{Nicolai2016} proposed a two-stream CNN architecture for frame-to-frame point cloud odometry estimation. They pre-processed point cloud data into 2-D images as inputs for their neural network. They collected point cloud data using the VLP-16 Lidar sensor mounted on Turtlebot. Although their experimental results are not superb, they demonstrate that it is possible to apply deep learning to LO task.

\section{Approaches for Point Cloud Encoding}
\begin{figure}
  \centering
  \includegraphics[width=\columnwidth]{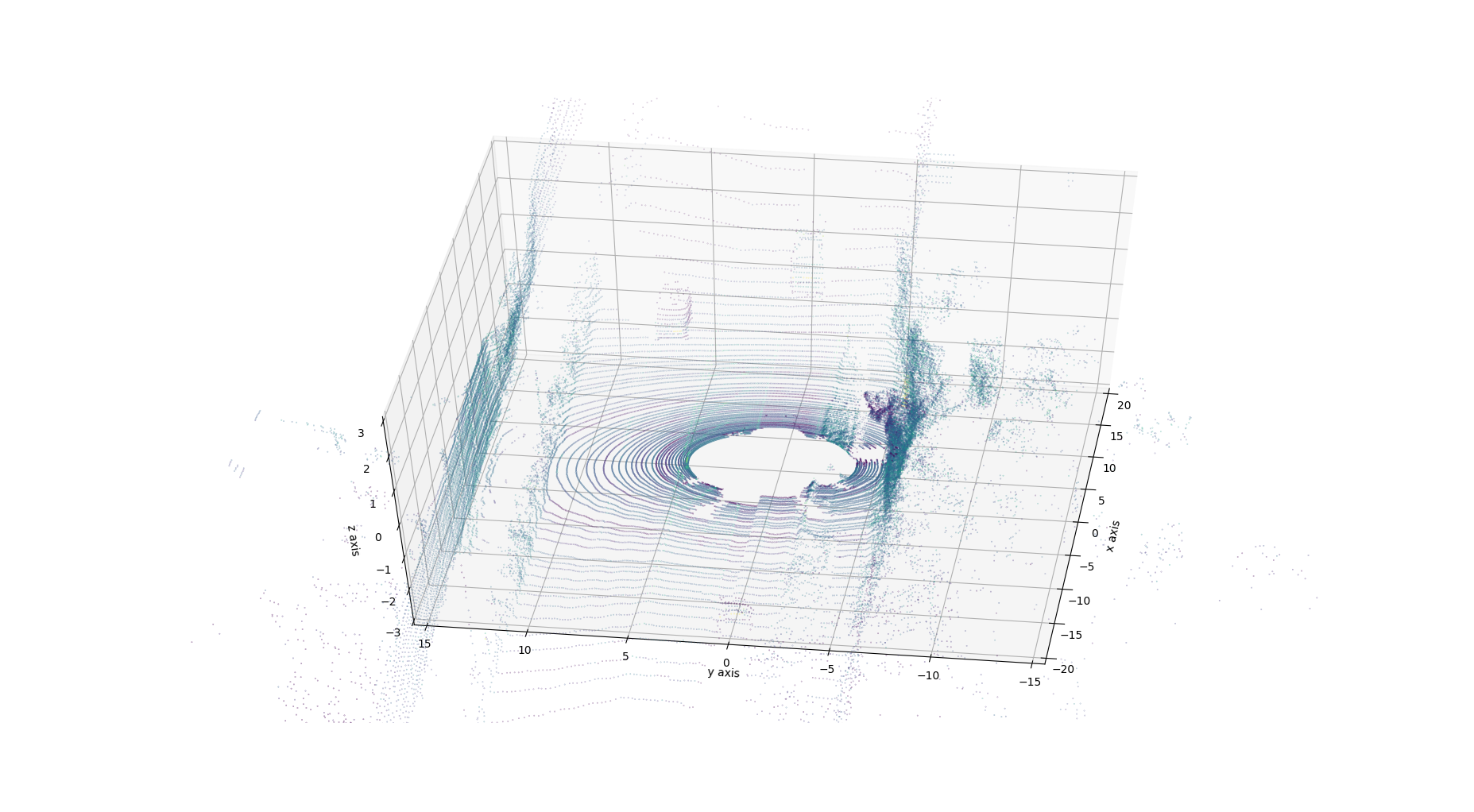}
  \caption{A sample point cloud from KITTI dataset.}
  \label{samplePointCloud}
\end{figure}

Convolutional neural networks require highly structured data as inputs whereas point cloud data is unordered and irregularly sampled. In this section, we will discuss different point cloud encoding approaches and compare their relative merits for the task of point cloud odometry. An example of point-cloud data is shown in Fig \ref{samplePointCloud}, as obtained by a car-mounted LIDAR. Note the irregular density and gaps due to obstructions.

\subsection{2-D Encoding of Point Clouds}\label{AA}
A straightforward approach to encode point clouds for odometry is to transform into a 2-D depth image. The projected depth image can be top-view, front-view, or panoramic-view. Among all of our experiments, we find that panoramic-view projection performs the best among all developed models with different parameter settings. Therefore, here we discuss panoramic-view projection equations that we have adopted in our work. Specifically, Li et al. \cite{Li} proposed an approach to project point cloud into panoramic depth image by the following equations:
\begin{align}
&\theta = \arctan\textit{2}(y, x) \\
&\phi = \arcsin(z/\sqrt{x^2 + y^2 + z^2}) \\ 
&r = \lfloor \theta / \Delta\theta \rfloor \\
&c = \lfloor \phi / \Delta\phi \rfloor
\end{align}
where (x,y,z) are 3-D point coordinates, $\theta$ is azimuth angle, $\phi$ is elevation angle, (r, c) is the 2-D image position of 3-D point projection, $\Delta\theta$ is average horizontal angle resolution and $\Delta\phi$ is vertical angle resolution. For a detailed explanation, we refer readers to the original paper. In our work, we used the same equations above, and then normalized depth values to the range [0, 255]. Points closer to the sensor are assigned higher values. Indeed, the inverse normalization has been successfully applied to many 2-D vision tasks.

\subsection{3-D Encoding of Point Clouds}
3-D point cloud data can also be discretized as a 3-D voxel grid and has been employed in VoxNet. Nevertheless, the voxelization of point clouds is impractical. For example, each point cloud in KITTI \cite{Geiger2012} contains approximately 120,000 points, which would require large memory size and computation resources to maintain the high resolution of the 3-D grid. Even adopting a sampling strategy to reduce the number of points, voxelization of point clouds is challenging for volumetric CNNs. Sampling would also affect the accuracy of our odometry task. Other than the voxelization approach, \cite{Qi2017} proposed PointNet, a straightforward approach that treats every single point as a 3-dimensional vector (x, y, z) for 2-D CNN architecture. Thus, the input points are encoded as $n \times 3$, and this approach eliminates the use of 3-D CNN architecture which greatly speeds up the training process. However, their task and other following works mainly deal with classification or segmentation \cite{Yang2019}. Moreover, the introduced transformation sub-network in the PointNet alters the translation that we need to accurately predict in our PCO task. Therefore, we reuse the PointNet architecture except we eliminate the T-Net in our experiment. More recently, \cite{Le2018} proposed a 3-D CNN architecture named PointGrid to integrate the benefits of point and grid for better representation of the local geometry shape given the limits of PointNet's ability to capture contextual neighborhood structure.

\begin{figure}[!t]
	\centering
	\subfigure[Original Point Cloud]{
	\begin{minipage}{0.45\columnwidth}
	\includegraphics[height=0.7\columnwidth]{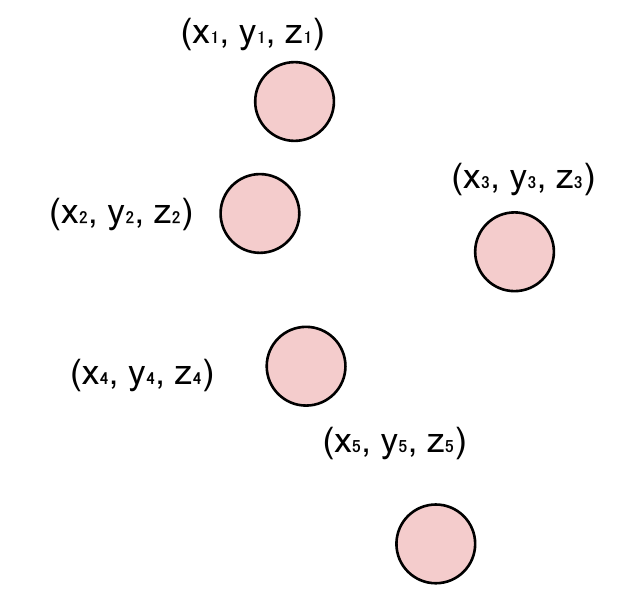} \\
	\end{minipage}
	}
	\subfigure[Panoramic Depth Image]{
	\begin{minipage}{0.45\columnwidth}
	\includegraphics[height=0.7\columnwidth]{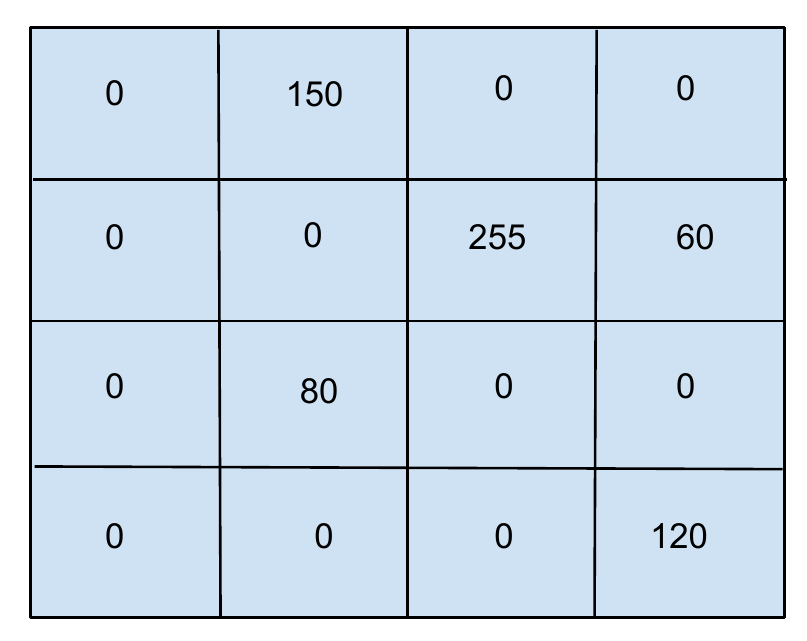} \\
	\end{minipage}
    }
    \subfigure[PointNet 3-D Encoding]{
	\begin{minipage}{0.45\columnwidth}
	\includegraphics[height=0.7\textwidth]{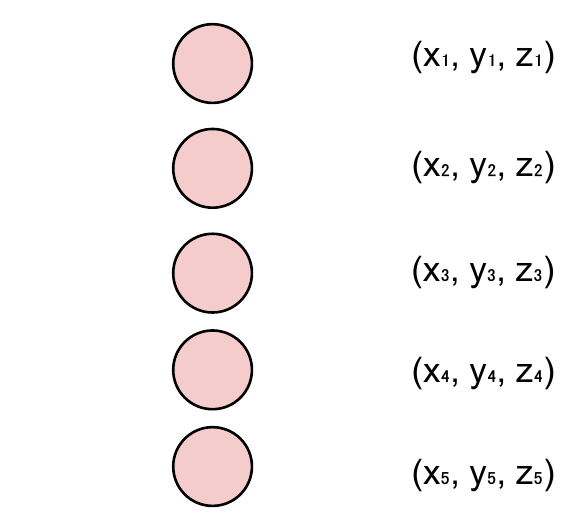} \\
	\end{minipage}
    }
	\subfigure[PointGrid 3-D Encoding]{
	\begin{minipage}{0.45\columnwidth}
	\includegraphics[height=0.7\textwidth]{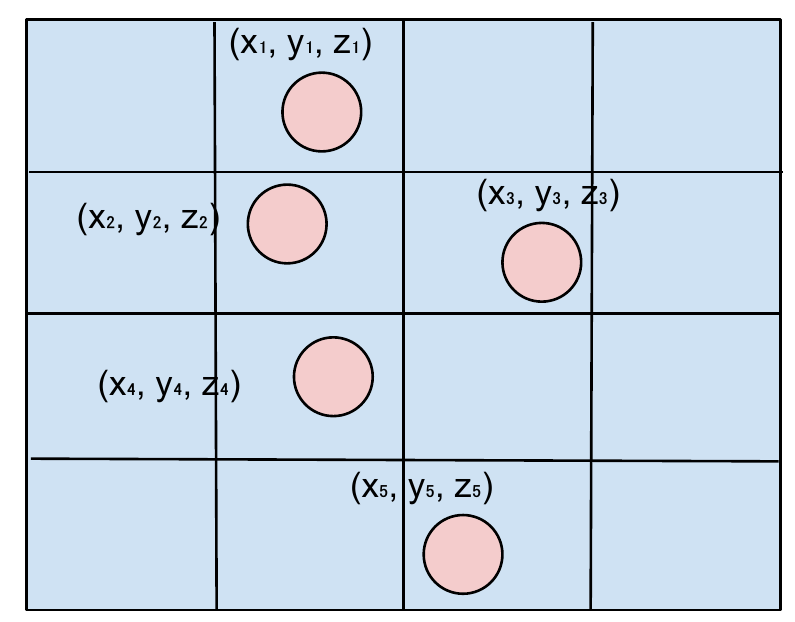} \\
	\end{minipage}
    }
    \caption{Visualization of various point cloud encoding approaches for deep learning.}
	\label{encoding}
\end{figure}

A graphical representation of all point cloud encoding approaches for deep learning in this work have been summarized in Fig.\ref{encoding}.
We used 2-D encoding for our proposed neural network architecture, termed DeepPCO, in which the point cloud data is projected to panoramic depth images. In the experimental results, we compare its performance with the PointNet and PointGrid 3-D encoding approaches.

\section{DeepPCO Architecture}
\begin{figure*}
  \centering
  \includegraphics[width=\linewidth]{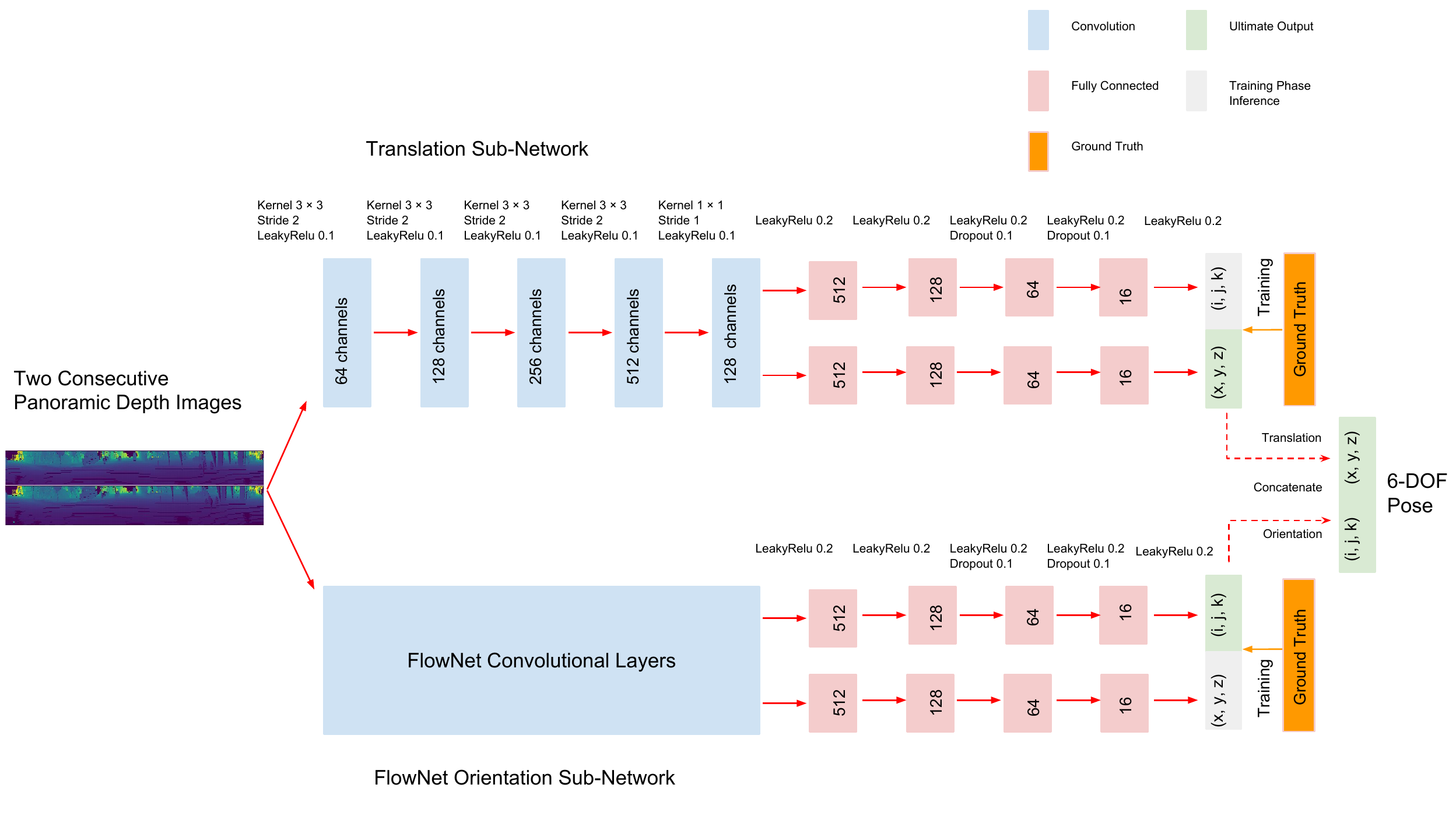}
  \caption{The architecture of DeepPCO. Two consecutive point clouds are encoded into panoramic depth images. These image pairs are sent to two Sub-Networks simultaneously, both of which are trained using 6-DOF pose ground truth. The Translation Sub-Network is used to predict translation vector (x, y, z) while the FlowNet Orientation Sub-Network infers the orientation vector (i, j, k). The translation and orientation vector are concatenated to output the desired 3-D transformation. All the initial configurations are listed, and we employ Dropout to prevent overfitting during training. Values shown in convolutional layers represent the number of channels/depth. }
  \label{deeppco_architecture}
\end{figure*}

Our DeepPCO is composed of two sub-networks, a Translation Sub-Network and FlowNet Orientation Sub-Network, both of which form a deep parallel neural network architecture. The entire DeepPCO architecture is illustrated in Fig. \ref{deeppco_architecture}.

\subsection{End-to-End Network Architecture}
Our key idea is that instead of jointly learning and predicting position and rotation vectors using a single neural network as adopted by existing work, we design two separate neural networks - one is adept at predicting translation while the other specializes in inferring rotation. As will be shown in the experimental results, this approach leads to superior overall performance. The inputs of DeepPCO are two consecutive 2-D panoramic-view projection images which are stacked together. The Translation Sub-Network is responsible for the prediction of translation between two point clouds, while the FlowNet Sub-Network infers the rotation between them. We used FlowNet as it has been proved to effectively extract geometrical features useful for odometry task. In Translation Sub-Network, fully convolutional layers are utilized to extract features from projected depth images, while in FlowNet Sub-Network, we adopt the same configuration of convolutional layers from FlowNet\cite{Fischer2015}. Our intuition to take advantage of FlowNet is that it is constructed upon optical flow, which is heavily used in VO tasks to attain good rotation performance. Since the number of input channels of FlowNet is six, we simply expand each depth image to three channels by replicating it twice. Leaky Rectified Linear Units (Leaky ReLU) are applied after every convolutional layer in our architecture with the angle of negative slope set to 0.1 and using in-place operation. We employ transfer learning to initialize the weights of convolutinal layers in our FlowNet Sub-Network using pretrained FlowNet convolutional layers' weights. Note that we do not use any max pooling layers or batch normalization layers in our architecture since our comprehensive experiments indicate that adding either of these two layers significantly decreases the prediction accuracy. Moreover, both of these two sub-networks have the same fully connected layers settings as in the \cite{Nicolai2016}, which consists of two branches jointly trained by 6-DOF ground truth. Although their work used their own collected indoor point cloud dataset to design the network, our experiment empirically proves that such a setting of fully connected layers has the best performance for PCO. Another important design is that we jointly learn poses for each sub-network. For example, for the Translation Sub-Network, although our final goal of this sub-network is to output translation vector, during the training phase, we jointly trained translation and orientation together as illustrated in Fig. \ref{deeppco_architecture} (orange color stands for 6-DOF pose). This is inspired by Grimes et al. \cite{Grimes2015}, which demonstrated that regressing position and orientation separately performed worse compared to the full 6-DOF pose for training. To summarize, our neural network architecture is end-to-end, and easy to be trained since it only employs a few 2-D CNN layers and a small fully connected layers, eliminating the use of large RCNN architecture.

\subsection{Output and Cost Function}
Our network outputs a 6-DOF pose vector \textbf{v}:
\begin{align}
\textbf{v} = \textbf{[p, q]}
\end{align}
where \textbf{p} is a 3-D position and \textbf{q} is an Euler angle. We express orientation of pose as Euler angle rather than quaternion based on the reason that quaternion is subject to an extra unit constraint affecting the optimization of our network. 
Since the two sub-networks are trained in a supervised manner simultaneously, our final objective loss function \textit{L} across two sub-networks is defined as follows:
\begin{align}
\textit{L} = \| \textbf{p} - \hat{\textbf{p}} \|_{2}^{2} + k * \| \textbf{q} - \hat{\textbf{q}} \|_{2}^{2}
\end{align}
where $\hat{\textbf{p}}$ is ground truth of translation vector, $\hat{\textbf{q}}$ is ground truth of orientation vector, $\| * \| _{2}^{2}$ measures the mean squared error (squared L2 norm) between each element in the estimation vector and ground truth vector, and $k$ is a scale factor to balance the errors of position and orientation to be approximately equal.

\section{Experimental Evaluation}
In this section, we introduce our experiment and training details. We compare our network to 5 baseline approaches, which are designed and tested on their best variations for our task. The \textbf{two-stream} network \cite{Nicolai2016} performs the same task as ours and is set as the benchmark for the PCO task. We modified their original network to infer 6-DoF pose, and we retrain it from scratch on the same input data as the DeepPCO. \textbf{DeepVO} and \textbf{ResNet18} \cite{He2016} are deep-learning approaches for visual odometry tasks and we use 2-D panoramic depth images as inputs. We select \textbf{PointNet} and \textbf{PointGrid} as comparisons for the 3-D encoding of point cloud due to their excellent performance in tasks such as semantic segmentation. For the PointNet, we remove the input transformation part.

\subsection{Dataset}
The KITTI VO/SLAM benchmark dataset is chosen for the experiment. The dataset consists of 11 sequences (Sequences 00-10), which are associated with ground truth pose. Each sequence contains consecutive point cloud data collected by a laser sensor for trajectory estimation. Since the dataset is relatively small compared to other benchmark datasets, we selected Sequence 00-03, 05-09 as training datasets, and Sequence 04 and 10 as test datasets which are chosen by most VO work since they can represent the major scenarios for agent or sensor movements in real world. Considering our loss function adopts Euler angle while the orientation of ground truth in KITTI is quaternion, we transformed orientation inside our experimental datasets to the Euler angle. All poses in KITTI are based on absolute transformation, so we converted them to relative transformation for the purpose of training the network.

\subsection{Implementation and Settings}
Our architecture is implemented with PyTorch framework. For evaluating our results, we choose Root Mean Square Error (RMSE) as our metric. Adam \cite{Kingma2015} optimizer was employed to train our neural network with $\beta_{1} = 0.9$ and $\beta_{2} = 0.999$. We set the learning rate from 0.0001 and then we applied 50\% weight decay for every 10 epochs. Our network was trained for 30 epochs according to different encoding approaches and architectures. Batch size was set to 8. The scale factor $k$ of the objective function is 100. We use 8-fold cross-validation for choosing hyperparameters. During the training phase, we also used data shuffling to prevent overfitting.

\subsection{Test Results}
The test results are presented in Table \ref{table:kitti}. Our experiment indicates that the proposed architecture outperforms all the baselines for both test sequences. A deeper convolutional neural network architecture, like ResNet18, is not necessary for achieving excellent odometry results. Among 3-D encoding approaches, PointGrid is more suitable for performing 6-DOF tasks than PointNet. In particular, our fully connected layers share very similar structure with the Two-stream approach. We trained them using the same settings on the same KITTI datasets, and the main difference is the convolutional parts, which we utilize more filters for feature extraction. Hence, it indicates that good feature extraction plays a significant role for odometry.

\begin{table}
\centering
\caption{\small Test results on KITTI sequences: $t_{rel}$ is averaged translation RMSE between prediction and ground truth, and $r_{rel}$ is averaged orientation RMSE between prediction and ground truth. The measurement units are \textbf{m}).}
\label{table:kitti}
\begin{tabular}{|c|c|c|c|c|}
\hline
&\multicolumn{2}{|c|}{\textbf{Sequence 04}}&\multicolumn{2}{|c|}{\textbf{Sequence 10}}\\
\hline
\textbf{Model}         & t\_{rel}    & r\_{rel}  & t\_{rel}    & r\_{rel} \\ \hline
Two-stream\cite{Nicolai2016}     & 0.0554     & 0.0830   & 0.0870     & 0.1592\\
ResNet18\cite{He2016}               & 0.1094     & 0.0602   & 0.1443     & 0.1327\\
DeepVO\cite{Wang2017}                 & 0.2157     & 0.0709   & 0.2153     & 0.3311\\
PointNet\cite{Qi2017}               & 0.0946     & 0.0442   & 0.1381     & 0.1360\\
PointGrid\cite{Le2018}              & 0.0550     & 0.0690   & 0.0842     & 0.1523\\ \hline
\textbf{DeepPCO (Ours)}          & \textbf{0.0263} & \textbf{0.0305} & \textbf{0.0247} & \textbf{0.0659}\\
\hline
\end{tabular}
\end{table}

\subsection{Trajectory Evaluation}
\begin{figure}[!t]
	\centering
	\subfigure[Sequence 04]{
	\begin{minipage}{0.225\textwidth}
	\includegraphics[width=1.0\textwidth]{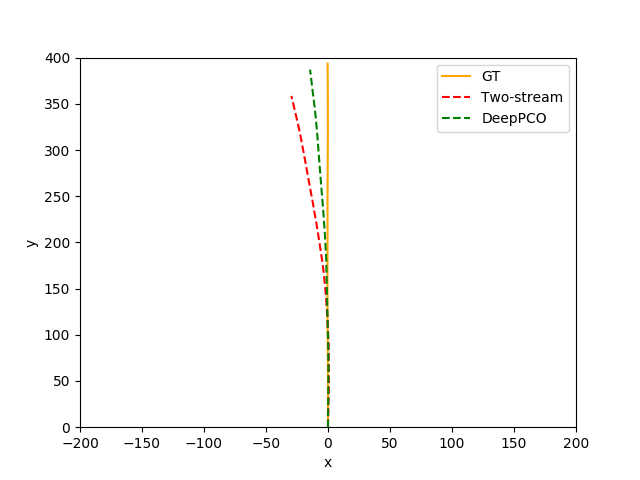} \\
	\end{minipage}
	}
	\subfigure[Sequence 10]{
	\begin{minipage}{0.225\textwidth}
	\includegraphics[width=1.0\textwidth]{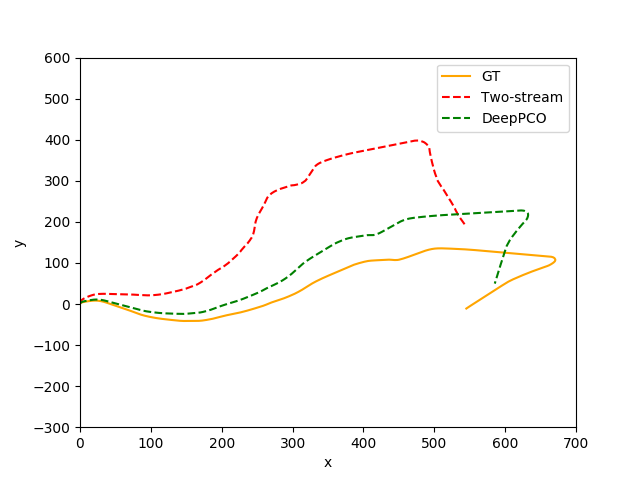} \\
	\end{minipage}
    }
    \caption{Trajectories of test results on Sequence 04 and 10.}
	\label{trajectory}
\end{figure}

Qualitative results are shown in Fig. \ref{trajectory} where we plot predicted trajectories of DeepPCO compared with TwoStream, as the most comparable baseline.

\begin{figure}
      \centering
      \includegraphics[width=\linewidth]{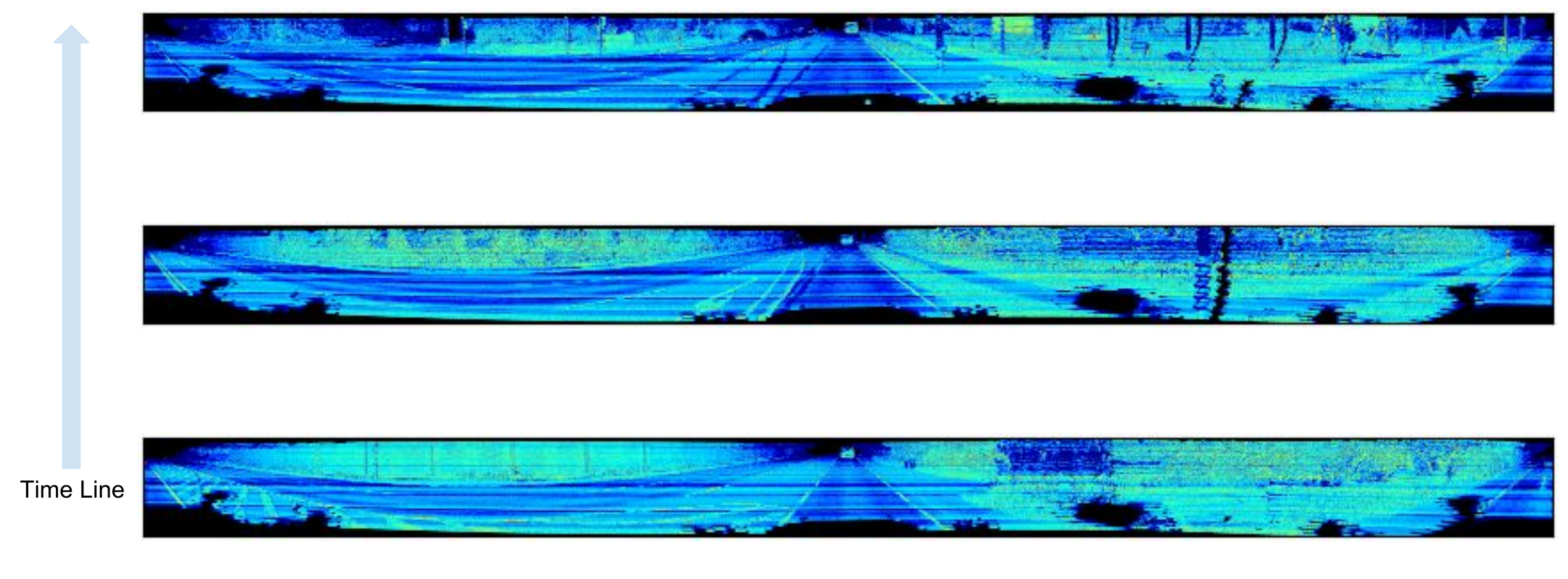}
      \caption{Sample projected depth images from sequence 04 in the latter timeline. Images are plotted using depth values from the range [0, 255]. The surroundings are empty and lack features.}
       \label{sequence04}
    \end{figure}

The proposed DeepPCO system can produce accurate pose estimation with respect to ground truth. Compared to the baseline approach, DeepPCO shows significant improvement. As seen in Fig. \ref{trajectory}, during the first 200 metres of Sequence 04, DeepPCO has very low drift. However, after 200 meters, the drift gradually increases. In order to investigate the reason, we plotted the projected depth images as shown in Fig. \ref{sequence04}.
The plot suggests that large open areas can degrade the accuracy of pose prediction to some degree. Combining with inertial measurements could overcome some of these issues.

\subsection{Comparison to Conventional Approaches}
In our experiments, we also compared our work to the conventional approach. We run open-sourced LOAM using point clouds. For a fair comparison, we did not use any IMU data, and we employed the official evaluation tool released for KITTI to examine these two approaches. We take sequence 04 as an example. The percentage of translation error and deg/m of rotation error of LOAM are 2.3245\% and 0.0108, while the results of DeepPCO are 3.1012\% and 0.0177 respectively. We notice that the conventional approach outperforms the proposed technique, but DeepPCO still achieves good performance. One reason may be that the training datasets are limited, especially for rotation-related data in the KITTI datasets. The other reason may be that we did not utilize any geometric awareness in our network, which could be further explored in the future.  

\subsection{Ablation Study}
In order to explore the impact of various components of our DeepPCO, we conduct the following ablation experiments. First, we split each sub-network to individually infer 6-DOF poses and examine whether our parallel architecture can outperform the individual ones. Second, for each sub-network, we keep the convolutional layers the same but use combined fully connected layers instead of two branches of fully connected layers in DeepPCO. Our purpose is to check whether the design of two branches for predicting translation and orientation separately is better than a single branch to jointly train and predict transformation. We note that all variants of our ablation architectures are trained from scratch. 
\begin{enumerate}[label=\alph*)]
	\item \textbf{\emph{Individual Sub-Network}}: Since we trained our sub-networks using full 6-DOF ground truth, we want to investigate whether any single sub-network was good enough to predict pose. Hence, we segmented our parallel architecture to two individual networks but kept all the parameters and hyperparameter settings unchanged. In order to ensure the fairness of our experiments, we randomly run 5 times for each sub-network to evaluate translation and orientation on sequence 04 and 10 respectively. Meanwhile, we calculated the average of RMSE for the 5 experiments. All the results are shown in Table \ref{table:individualSubNetwork}. Results suggest that all translation predictions on Sequence 04 and 10 using Translation Sub-Network are better than FlowNet Orientation Sub-Network, while all orientation predictions based on FlowNet Orientation Sub-Network are better than those generating from Translation Sub-Network. In other words, this ablation experiment demonstrated that Translation Sub-Network performs better for inferring translation whereas achieving worse orientation prediction than FlowNet Orientation Sub-Network. Exactly the opposite, FlowNet Orientation Sub-Network outperforms Translation Sub-Network for orientation inference. Therefore, it is a desirable choice for designing dual sub-networks to predict translation and orientation separately.
	
\begin{table*}
\centering
\caption{\small Test results of individual sub-network on KITTI sequences: $t_{rel}$ is averaged translation RMSE between prediction and ground truth, and $r_{rel}$ is averaged orientation RMSE between prediction and ground truth. The measurement unit is meter (\textbf{m}). Each sub-network was trained using 6-DOF poses, and we conducted 5 times independently for each one.}
\label{table:individualSubNetwork}
\begin{tabular}{|c|c|c|c|c|c|c|c|c|}
\hline
&\multicolumn{4}{|c|}{\textbf{Translation}}&\multicolumn{4}{|c|}{\textbf{FlowNet Orientation}}\\
&\multicolumn{4}{|c|}{\textbf{Sub-Network Only}}&\multicolumn{4}{|c|}{\textbf{Sub-Network Only}}\\
\hline
&\multicolumn{2}{|c|}{\textbf{Sequence 04}}&\multicolumn{2}{|c|}{\textbf{Sequence 10}}&\multicolumn{2}{|c|}{\textbf{Sequence 04}}&\multicolumn{2}{|c|}{\textbf{Sequence 10}}\\
\hline
\textbf{Times}         & t\_{rel}    & r\_{rel}  & t\_{rel}    & r\_{rel} & t\_{rel}    & r\_{rel}  & t\_{rel}    & r\_{rel}\\ \hline
1     & \textbf{0.0288}     & 0.0421   & \textbf{0.0243}    & 0.0735      & 0.0474     & \textbf{0.0315}   & 0.0352     & \textbf{0.0648} \\
2     & \textbf{0.0282}     & 0.0392   & \textbf{0.0235}    & 0.0731      & 0.0471     & \textbf{0.0312}   & 0.0373     & \textbf{0.0675} \\
3     & \textbf{0.0262}     & 0.0429   & \textbf{0.0248}    & 0.0737      & 0.0469     & \textbf{0.0297}   & 0.0377     & \textbf{0.0682} \\
4     & \textbf{0.0307}     & 0.0417   & \textbf{0.0234}    & 0.0755      & 0.0469     & \textbf{0.0302}   & 0.0342     & \textbf{0.0676} \\
5     & \textbf{0.0281}     & 0.0426   & \textbf{0.0247}    & 0.0740      & 0.0580     & \textbf{0.0309}   & 0.0447     & \textbf{0.0669} \\ \hline
\textbf{Mean}  & \textbf{0.0284} & 0.0417 & \textbf{0.0241} & 0.0740    & 0.0493 & \textbf{0.0307} & 0.0378 & \textbf{0.0670}\\
\hline
\end{tabular}
\end{table*}
	
	\item \textbf{\emph{Single Branch Fully Connected Layers}} In our proposed architecture, we utilized two branches for predicting translation and orientation. However, we consider whether only using a single branch of fully connected layers if each sub-network can produce a better result. Thus, we design the ablation study for this situation as shown in Fig. \ref{singleBranch}. 
	\begin{figure}
      \centering
      \includegraphics[width=\linewidth]{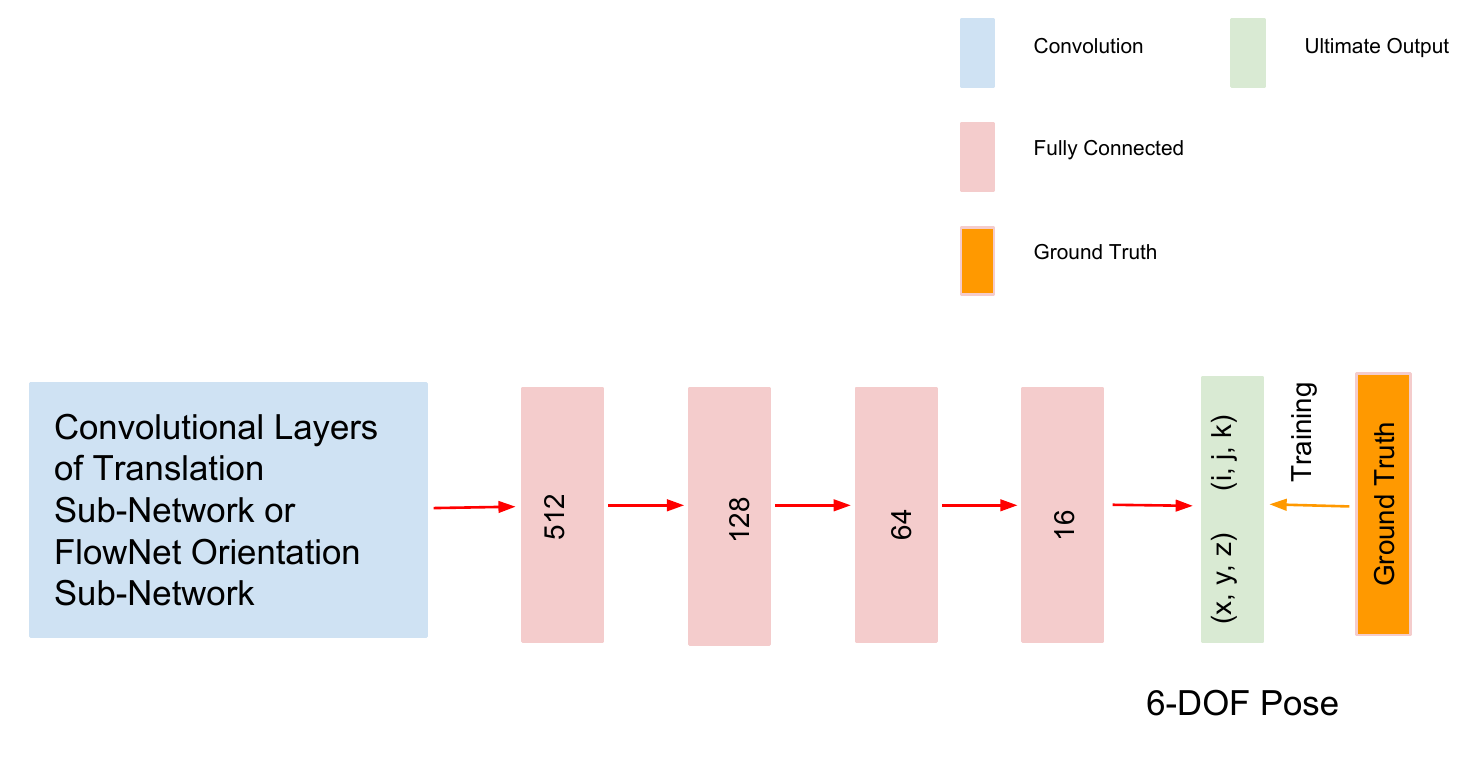}
      \caption{Ablation experiment of single branch fully connected layers. All the parameter configurations of convolutional layers and fully connected layers are the same as DeepPCO. Different from DeepPCO in which transformation vector is trained using two branches, 3-D translation (x, y, z) and orientation (i, j, k) are jointly trained and inferred by just one branch here.}
        \label{singleBranch}
    \end{figure}
	All the parameter configurations of layers remain the same. The only difference is that we use one branch to jointly learn transformation. The results are presented in Table \ref{table:singleBranch}. While single branch FC layers are capable of attaining reasonable estimations, they are less accurate than our parallel architecture. Results indicate that adopting the design of two branches of FC layers in DeepPCO tends to have better transformation inference for the PCO task.

\begin{table}
\centering
\caption{\small Test results of various branches of fully connected (FC) layers on KITTI sequences: $t_{rel}$ is averaged translation RMSE between prediction and ground truth, and $r_{rel}$ is averaged orientation RMSE between prediction and ground truth. The measurement unit is meter (\textbf{m}). For single branch FC layers, each sub-network was trained using full 6-DOF pose while for DeepVO, each sub-network was learned by two branches, one for translation and the other for orientation.}
\label{table:singleBranch}
\begin{tabular}{|c|c|c|c|c|}
\hline
&\multicolumn{2}{|c|}{\textbf{Sequence 04}}&\multicolumn{2}{|c|}{\textbf{Sequence 10}}\\
\hline
\textbf{Model}         & t\_{rel}    & r\_{rel}  & t\_{rel}    & r\_{rel} \\ \hline
Single Branch FC     & 0.0559     & 0.0843   & 0.0327     & 0.0729 \\ \hline
Two Branches FC     & \textbf{0.0284}     & \textbf{0.0307}   & \textbf{0.0241}     & \textbf{0.0670} \\
\hline
\end{tabular}
\end{table}
\end{enumerate}

\section{Conclusion}
In this paper, we have presented an end-to-end deep parallel neural network named DeepPCO for the point cloud odometry task. Two consecutive point clouds are processed into panoramic depth images, which are stacked and sent simultaneously to dual sub-networks for transformation estimation. Our approach shows good performance for the odometry task in the 3-D real-world environment. Several interesting extensions could be considered from our work, such as integrating our system to a full learning-based SLAM system, evaluating our approach on more challenging environments like severe weather, or developing learning-based sensor fusion methods for odometry tasks. In the future, we plan to add geometric optimization to our architecture, which may further improve our performance. We hope our design choices and experiments on 2-D and 3-D encoding approaches and corresponded architectures can inspire further research for point cloud-related tasks and system developments. We believe that this work is an important step towards developing robust point cloud-based odometry.

\section*{Acknowledgment}
This work is funded by the NIST grant ``Pervasive, Accurate, and Reliable Location-Based Services for Emergency Responders''. The authors would like to thank Linhai Xie, Stefano Rosa, Sen Wang and Ronnie Clark for the fruitful discussion and suggestions.

\addtolength{\textheight}{-12cm}   

\bibliographystyle{abbrv}
\bibliography{references}

\end{document}